\documentclass[conference]{IEEEtran}
\IEEEoverridecommandlockouts
\usepackage{cite}
\usepackage{amsmath,amssymb,amsfonts}
\usepackage{algorithmic}
\usepackage{graphicx}
\usepackage{textcomp}
\usepackage{xcolor}
\usepackage{listings}
\usepackage{tcolorbox}
\usepackage{hyperref}
\usepackage[]{mdframed}
\usepackage[hybrid]{markdown}
\usepackage{array}
\usepackage{booktabs}
\usepackage{newtxtext}

\def\BibTeX{{\rm B\kern-.05em{\sc i\kern-.025em b}\kern-.08em
    T\kern-.1667em\lower.7ex\hbox{E}\kern-.125emX}}
\begin{document}

\newcolumntype{L}[1]{>{\raggedright\arraybackslash}p{#1}}
\newcolumntype{C}[1]{>{\centering\arraybackslash}p{#1}}
\newcolumntype{R}[1]{>{\raggedleft\arraybackslash}p{#1}}

\lstdefinestyle{mystyle}{
  basicstyle=\ttfamily\small,
  columns=fullflexible,
  breaklines=true,
  frame=none,
  xleftmargin=2em,
  xrightmargin=2em
}

\lstset{style=mystyle}

\newcommand{\custombox}[2]{
  \begin{tcolorbox}[colback=white, colframe=black, fonttitle=\bfseries, title=#1]
    \lstinline|#2|
  \end{tcolorbox}
}

\lstdefinelanguage{json}{
  basicstyle=\ttfamily,
  showstringspaces=false,
  breaklines=true,
  literate=
   *{0}{{{\color{numb}0}}}{1}
    {1}{{{\color{numb}1}}}{1}
    {2}{{{\color{numb}2}}}{1}
    {3}{{{\color{numb}3}}}{1}
    {4}{{{\color{numb}4}}}{1}
    {5}{{{\color{numb}5}}}{1}
    {6}{{{\color{numb}6}}}{1}
    {7}{{{\color{numb}7}}}{1}
    {8}{{{\color{numb}8}}}{1}
    {9}{{{\color{numb}9}}}{1}
}

\setlength{\skip\footins}{8pt}

\title{ConvoGen: Enhancing Conversational AI with Synthetic Data: A Multi-Agent Approach}

\author{
\IEEEauthorblockN{Reem Gody}
\IEEEauthorblockA{\textit{Microsoft AI} \\
regody@microsoft.com}
\and
\IEEEauthorblockN{Mahmoud Goudy}
\IEEEauthorblockA{\textit{Microsoft AI} \\
mahmoudgoudy@microsoft.com}
\and
\IEEEauthorblockN{Ahmed Y. Tawfik}
\IEEEauthorblockA{\textit{Microsoft AI} \\
atawfik@microsoft.com}
}
\maketitle

\begin{abstract}
In this paper, we present ConvoGen: an innovative framework for generating synthetic conversational data using multi-agent systems. Our method leverages few-shot learning and introduces iterative sampling from a dynamically updated few-shot hub to create diverse and realistic conversational scenarios. The generated data has numerous applications, including training and evaluating conversational AI models, and augmenting existing datasets for tasks like conversational intent classification or conversation summarization. Our experiments demonstrate the effectiveness of this method in producing high-quality diverse synthetic conversational data, highlighting its potential to enhance the development and evaluation of conversational AI systems.
\end{abstract}


\section{Introduction}

Recent advancements in conversational AI have significantly enhanced the integration of AI across various applications, thereby improving the overall user experience. These advancements span from virtual assistants powered by large language models (LLMs), to video conferencing applications that leverage LLMs to generate summaries and insights from meeting conversations, and messaging or email applications that utilize AI for rewriting, or generating smart responses. These applications can either involve interactions between an agent and a user, or they may encompass open-domain social multi-party conversations. Accordingly, open-domain conversational data is a crucial enabler for such applications, driving research into different approaches for their collection.

One common approach is web crawling, exemplified by the DailyDialog dataset \cite{li2017DailyDialogmanuallylabelledmultiturn}, which was sourced from English learning websites. Another technique involves automatically creating conversations by scraping chat forums like Reddit \cite{mazare-etal-2018-training}, \cite{huryn-etal-2022-automatic}. Although scalable, this method often requires significant filtering to remove noise and toxic content.

Crowdsourcing is another prevalent methodology for collecting conversational data. Zhang et al. \cite{zhang-etal-2018-personalizing} introduced the PERSONA-CHAT dataset by assigning personas to crowd workers, resulting in more diverse and engaging dialogues. Similarly, Rashkin et al. \cite{rashkin2019empatheticopendomainconversationmodels} collected conversations by having one annotator describe an emotion-triggering situation and another react to it. While crowdsourcing can yield natural conversations, it is often costly, time-consuming, and challenging to scale for large datasets or multi-party conversations. Existing multi-party conversation datasets are typically derived from scripted TV series \cite{chen-choi-2016-character}, movies \cite{lison2016opensubtitles2016}, or speech corpora like the AMI meeting corpus \cite{kraaij2005ami}.

The generation of synthetic conversational data has become a focal point of research to augment existing datasets or create tailored data for specific scenarios \cite{abbasiantaeb2024let}, \cite{abdullin2024synthetic}, \cite{das2024synthetic}. Large Language Models (LLMs) have been extensively researched for their ability to generate synthetic conversational data for open-domain bots \cite{lee-etal-2022-personachatgen}, \cite{kim-etal-2023-soda}, \cite{chen-etal-2023-places}, \cite{jandaghi-etal-2024-faithful} and simulate human communication for applications like speech recognition \cite{zhang2024speechagentshumancommunicationsimulationmultimodal}, \cite{myat2024framework}.

A common strategy to diversify generated conversations is grounding them in common sense knowledge \cite{kim-etal-2023-soda}, specific topics \cite{chen-etal-2023-places}, or personas \cite{lee-etal-2022-personachatgen}, \cite{jandaghi-etal-2024-faithful}. For instance, Kim et al. \cite{kim-etal-2023-soda} used common sense knowledge from a knowledge graph to generate diverse dyadic conversations with GPT-3.5, suggesting their approach could extend to multi-party conversations. Chen et al. \cite{chen-etal-2023-places} grounded conversations in topics and backgrounds, generating both dyadic and triadic dialogues. Lee et al. \cite{lee-etal-2022-personachatgen} and Jandaghi et al. \cite{jandaghi-etal-2024-faithful} used personas to generate dyadic conversations, with the latter employing a generator-critic framework to ensure high-quality dialogues.

Our research introduces ConvoGen \footnote{\href{https://github.com/phantomcoder1996/autogen/tree/addconvogen/python/samples/convogen\_syntheticdatagen}{ConvoGen Link}}: an innovative multi-agent framework for generating diverse dyadic and multi-party conversations. Similar to previous works \cite{lee-etal-2022-personachatgen}, \cite{jandaghi-etal-2024-faithful}, \cite{zhang2024speechagentshumancommunicationsimulationmultimodal}, \cite{myat2024framework}, we utilize a persona-based approach to initiate group conversations between multiple agents. However, we employ an innovative approach to enhance the diversity of the generated conversations. Our primary contributions are as follows:
\begin{enumerate}

\item \textbf{Multi-Agent Framework for Synthetic Conversation Generation:} We propose a multi-agent framework as an intuitive solution to address the problem of generating multi-party conversational datasets involving two or more participants in an efficient and scalable way compared to traditional crowd-sourcing methods. The framework leverages LLM-generated experiences involving personas with specific relationships discussing a topic. By utilizing persona-based agents, we achieve diverse and varied conversations backed by the unique interactions between the matched personas that are likely to improve alongside advancements in persona generation \cite{ge2024scaling}.

\item \textbf{Iterative Sampling Technique:} We introduce iterative sampling as a technique to enhance the LLM-based experience generator and increase the diversity in the conversations generated based on these experiences in the case of large-scale dataset generation. Our evaluation demonstrates that iterative sampling results in high lexical diversity (average Measure of Lexical Diversity (MTLD) score), validating its effectiveness in generating varied conversational content.


\item \textbf{Extensive Evaluation:} We conduct extensive evaluations of the datasets generated by our framework to assess the lexical diversity and groundedness of the generated conversations across various setups. Using the Measure of Lexical Diversity (MTLD) metric, we demonstrate that LLM-based agents produce conversations with higher lexical diversity compared to casual human conversational datasets. Additionally, our LLM-as-a-judge approach validates that the generated conversations are well-grounded in the input personas, and the situations in which they are involved. 
\end{enumerate}

\begin{figure}[htbp]
\centerline{\includegraphics[width=\linewidth]{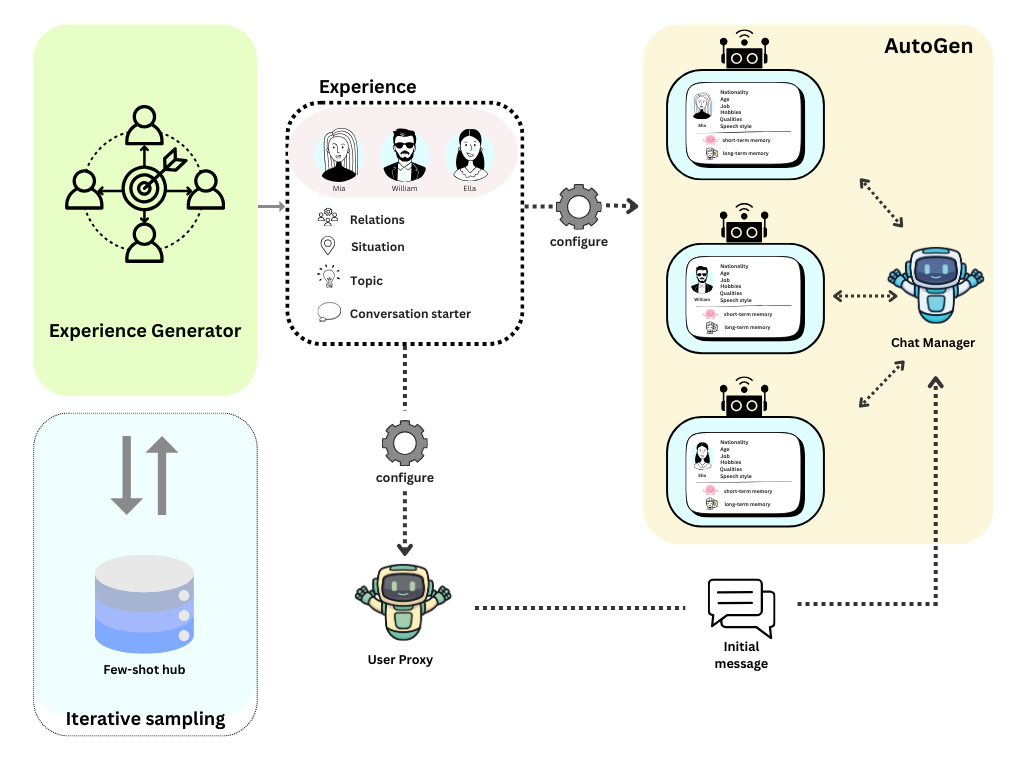}}
\caption{The procedure for generating conversations using ConvoGen. First, the experience generator generates the personas, their relations, a situation, a topic and a conversation starter. Next, a group of agents initialized using these personas engage in a conversation using the generated conversation starter. }
\label{fig:multiagent}
\end{figure}

\section{Methodology}
\label{section:methodology}
We develop a multi-agent based approach for general multi-party conversational data generation. In our approach, we use the AutoGen framework \cite{Wu_AutoGen_Enabling_Next-Gen_2024} to build our multi-agent system. We first give a brief overview of multi-agent systems and the AutoGen framework, and then we describe our approach for synthetic data generation. 
\subsection{Multi-agent Systems}
LLMs have demonstrated impressive capabilities in performing tasks, when configured with task-specific prompts, or assisted with access to external tools or data stores. In addition, it has been shown that LLMs are able to solve complex tasks when these tasks are broken down into simpler subtasks \cite{Guo2024LargeLM}. This has encouraged exploring the potential of building systems that involve multiple LLM agents, each specialized in doing a specific task, and allowing them to work together in a collaborative pattern to accomplish even more complex tasks. Accordingly, several frameworks have emerged to facilitate utilizing the multi-agent approach in building LLM applications such as CrewAI and AutoGen. These frameworks facilitate the integration and coordination of multiple LLM agents, enabling them to work seamlessly together to solve complex problems.
\subsection{AutoGen Framework}
AutoGen is an open-source framework designed to build multi-agent conversational systems using LLMs. It supports asynchronous messaging and both event-driven and request-response interaction patterns. The framework implements advanced multi-agent design patterns, including reflection, handoffs, mixture of agents, multi-agent debate, and group chat. In the group chat pattern, agents communicate simultaneously, taking turns to publish messages, with a group chat manager maintaining the order of turns. The manager selects the next speaker using either a round-robin algorithm or an LLM-based mechanism, then broadcasts the selected speaker's message to all agents. This renders group chat pattern an attractive and intuitive option for generating open-domain multi-party synthetic conversational data.

\subsection{Approach}
\begin{figure}[h!]
    \centering
    \includegraphics[width=0.75\linewidth]{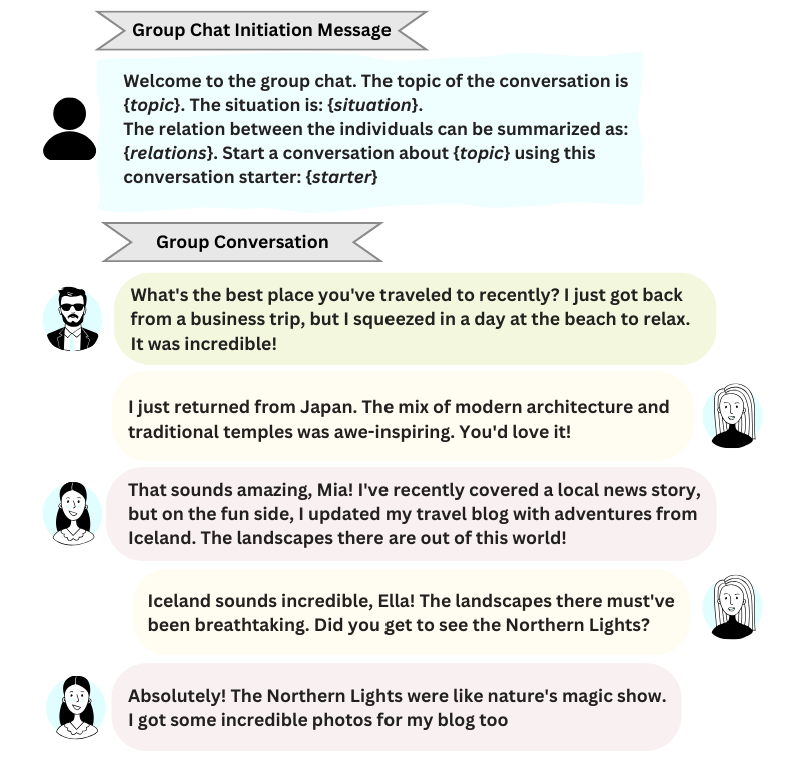}
    \caption{The message that is sent by the user proxy to the chat manager to initiate the group conversation, and an example of a generated group chat conversation. }
    \label{fig:initiator_message_conversation}
\end{figure}

We develop ConvoGen: a framework for synthetic data generation that leverages the AutoGen framework to instantiate group conversations between multiple pre-configured agents. Figure \ref{fig:multiagent} shows the architecture used for the generation of synthetic conversational data using multi-agent systems. Our approach is composed of two main steps:
\begin{enumerate}
\item \textbf{Experience Generation}

\noindent \hspace{0.5em} The primary intuition behind this step is that individuals' backgrounds, relationships, and personalities significantly influence their interactions. Additionally, the overall context and setting of a conversation play a crucial role in shaping how individuals drive the dialogue. To this end, we develop an experience generator powered by GPT-4o, leveraging few-shot learning to create diverse experiences. An experience is defined as a group of related personas and a situation that brings them together. This group discusses a specific topic using a designated conversation starter.

\noindent \hspace{0.5em} To achieve this, we experiment with two methods. In the first method, we provide guidelines in the LLM instructions to define the number of participants and their personas. Each persona is characterized by a name, qualities (e.g., friendly, humorous), lifestyle (work and hobbies), and speech style. Additionally, each persona has a memory, that is influenced by recent and long-term experiences. 

\noindent \hspace{0.5em} Subsequently, the LLM receives instructions to define the relationships between the participants and a situation influenced by the personas' memories. The LLM then generates a topic influenced by the situation and a conversation starter for that topic. A list of example experiences is provided to the LLM as shots to guide the generation process. In our experiments, we limit this list to one example to conserve context tokens. Figure \ref{fig:experience_personas} illustrates an example of an experience with a set of generated personas. 
Prompt \ref{prompt:experiencegen1} in Appendix \ref{appendix:genprompts} details these instructions. For the experience generator, we use a prompt that contains a placeholder for the shots that is dynamically replaced by specific examples during runtime. This allows us to sample experiences from a few-shot hub that is updated with the generated experiences during runtime. We refer to this approach as “iterative sampling" because the LLM iteratively updates the few-shot hub with the experiences it generates and learns from the shots it generates in the previous turns. This differs from dynamic few shot selection which focuses on selecting the most relevant examples to the task or user input.

\noindent \hspace{0.5em} The second method for generating experiences is similar to the first, with the key difference being that the prompt receives a list of predefined personas sampled from a persona hub during runtime. This is shown in prompt \ref{prompt:experiencegenpersona} in Appendix \ref{appendix:genprompts}. We experiment with the personas in \cite{zhang-etal-2018-personalizing} and \cite{ge2024scaling}. These personas have diverse backgrounds, leading to engaging conversations on various topics.

\begin{figure}[h!]
\centering
    \includegraphics[width=0.75\linewidth]{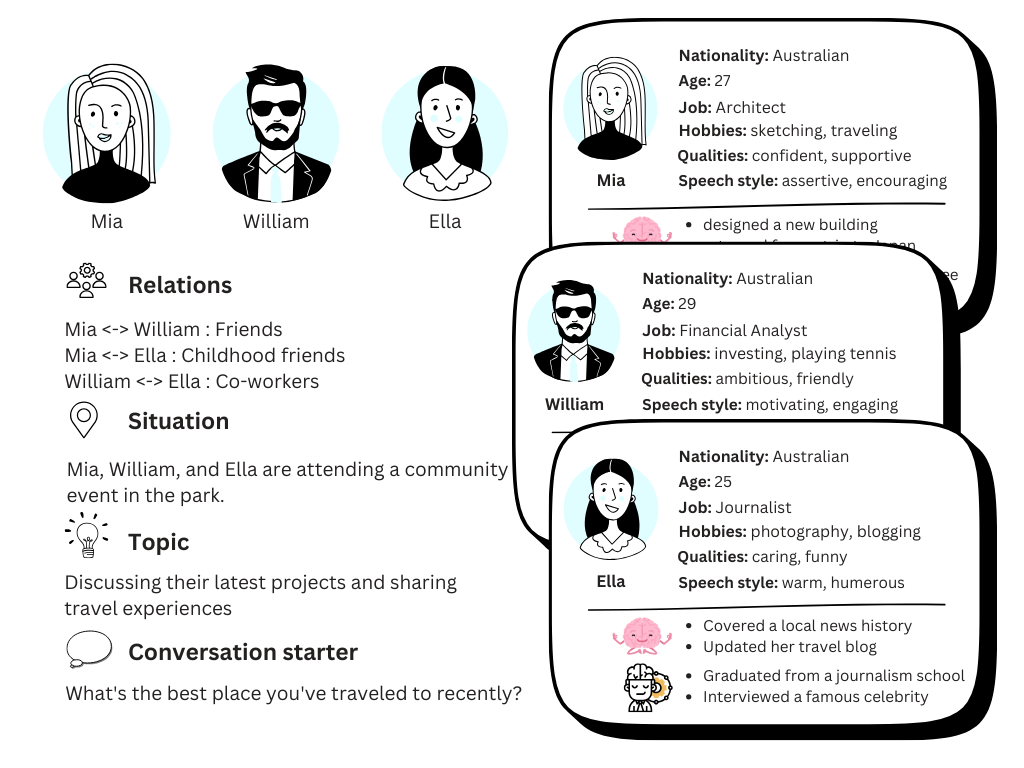}
    \caption{An example experience generated by the experience generator powered by the GPT-4o model. The experience involves a group of related personas, a topic and a conversation starter. }
    \label{fig:experience_personas}
\end{figure}

\item \textbf{Group Chat Instantiation}

\noindent \hspace{0.5em} We utilize the AutoGen framework for creating the group chat. As shown in Figure \ref{fig:multiagent}, the generated experiences are used to instantiate a group conversation between the multiple agents. First, each individual's persona is used to configure a corresponding agent using a system message, a name and a description. In addition to the persona definition, the system message includes additional guidelines for the agent on how to drive the conversation. 
These guidelines are detailed in Figure \ref{fig:agent-guidelines} in Appendix \ref{appendix:genprompts}. 
In our experiments, the guideline limiting the number of tokens per agent's response is very important to prevent the agents from being chatty and to generate more natural conversations.  Next, the conversation is instantiated by a user proxy who sends a message to the group chat manager composed of the situation, the relations between the individuals, the topic, and the conversation starter. The group chat manager then uses the predefined speaker selection prompt to select the next speaker from the list of agents based on the current conversation context and the agents' names and descriptions. The speaker responds to the group chat manager which hence broadcasts the message to all the other individuals in the group, and selects the next speaker to proceed. This process continues until a maximum number of turns is reached. Figure \ref{fig:initiator_message_conversation} demonstrates the initial message sent by the user proxy to trigger the group conversation, and an example of a group chat conversation generated by our framework.

\end{enumerate}

\vspace{-3mm}
\section{Experimental Setup}
\begin{table*}[tb]
    \centering
    \caption{Summary of all Experiments conducted with ConvoGen}
    \label{tab:experiments}
    \begin{tabular}{L{4cm}R{1.8cm}R{1.8cm}R{1.8cm}R{1.8cm}R{1.8cm}R{1.8cm}}
    \toprule
    Experiment & Abbreviation & nConv & Personas & Source & Iterative Sampling ? & Chatty ? \\
    \midrule
    Gen-Persona-Fix-Shot & GPFS & 504 & Generated & - & \text{\sffamily X} & \checkmark  \\
    Gen-Persona-Iterative-Sample & GPIS & 500 & Generated & - & \checkmark  & \checkmark  \\
    Sample-Persona-Fix-Shot & SPFS & 505 & Sampled & Billion Personas & \text{\sffamily X} & \checkmark  \\
    Sample-Persona-Iterative-Sample & SPIS & 491 & Sampled & Billion Personas & \checkmark  & \checkmark  \\
    \midrule
    Gen-Persona-Fix-Shot-Short & GPFS-s & 502 & Generated & - & \text{\sffamily X} & \text{\sffamily X} \\
    Gen-Persona-Iterative-Sample-Short & GPIS-s & 503 & Generated & - & \checkmark & \text{\sffamily X} \\
    Sample-Persona-Fix-Shot-Short & SPFS-s & 506 & Sampled & Billion Personas & \text{\sffamily X} & \text{\sffamily X} \\
    Sample-Persona-Iterative-Sample-Short & SPIS-s & 507 & Sampled & Billion Personas & \checkmark  & \text{\sffamily X} \\
    \midrule
    PERSONA-CHAT-Fix-Shot-Short & PCFS-s & 957 & Reused & PERSONA-CHAT & \text{\sffamily X} & \text{\sffamily X} \\
    PERSONA-CHAT-Iterative-Sample-Short & PCIS-s & 967 & Reused & PERSONA-CHAT & \checkmark  & \text{\sffamily X} \\
    \bottomrule
    \end{tabular}
\end{table*}
\label{section:experiments}
We conduct various experiments with our proposed methodology to generate conversational data. Each experiment configures the experience generator differently, and we analyze the datasets for lexical diversity and groundedness. We compare the lexical diversity results with human conversational datasets from Table \ref{tab:human} as a baseline. Typically, we generate around 500 conversations per experiment, except for PERSONA-CHAT based experiments, which match the number of conversations in the PERSONA-CHAT test set.
\begin{table}[b]
    \centering
    \caption{Summary of the human dataset baselines used, showing mainly the subset(s) and the total number of conversations}
    \label{tab:human}
    \begin{tabular}{L{3cm}R{2cm}R{1.7cm}}
    \toprule
    Dataset & Subset & nConv \\
    \midrule
    DailyDialog & train|valid|test & 13118 \\
    EmpatheticDialogues & train|valid|test & 24843 \\
    PERSONA-CHAT & test & 967 \\
    \bottomrule
    \end{tabular}
\end{table}
Following the same methodology described in section \ref{section:methodology}, the experience generator creates eight experiences concurrently. Each experience triggers a conversation using the AutoGen framework, resulting in eight conversations per run using the same shot.

Table \ref{tab:experiments} summarizes the experiments conducted. Personas are either \textbf{Generated}, \textbf{Sampled} from a large dataset, or \textbf{Reused} as is from existing datasets. Each configuration is tested once with a fixed shot for the experience generator and once using \textbf{Iterative Sampling}. In the first four experiments, agents tended to be too \textbf{chatty}, leading to long conversations. For the remaining experiments, we prompt the agents to be less chatty by explicitly adding an instruction for limiting the number of tokens per agent's response to achieve more normal interactions.

\vspace{-4mm}
\section{Analysis}
\label{section:analysis}
In this section, we analyze the conversational datasets produced as outlined in section \ref{section:experiments}. Initially, we present general statistics to compare these generated datasets. Subsequently, we evaluate the lexical diversity utilizing the Measure of Lexical Diversity (MTLD) metric \cite{McCarthy2010MTLDVA}, which helps gauge the richness and complexity of the generated conversations. Finally, we employ LLM evaluation based on a concept akin to the G-Eval framework \cite{Liu2023GEvalNE} to assess the alignment of the generated conversations with their input experiences produced by the experience generator.

\subsection{Statistics}
Table \ref{tab:statistics} compares statistics for baseline human datasets and our generated conversational datasets. Human conversations are less chatty, with fewer words per turn and shorter durations than synthetic ones. In the initial four experiments (\textbf{GPFS}, \textbf{GPIS}, \textbf{SPFS}, \textbf{SPIS}), agents produced excessively long conversations. After tuning prompts in the next four experiments (\textbf{GPFS-s}, \textbf{GPIS-s}, \textbf{SPFS-s}, \textbf{SPIS-s}), conversations became more natural, with fewer words per turn and shorter lengths.

\begin{table}[htbp]
    \caption{General Statistics for the generated conversations and human baselines, the reported numbers are the average number of words (nW) and the average number of turns (nT) per conversation along with the average number of words per turn nW/T}
    \centering
    \begin{tabular}{L{3cm}R{1.1cm}R{1.1cm}R{1.1cm}}
    \toprule
    Dataset & nW & nT & nW/T \\

    \midrule
    DailyDialog & 114.70 & 7.85 & 13.61 \\
    EmpatheticDialogues & 63.47 & 4.31 & 13.71 \\
    PERSONA-CHAT & 198.46 & 15.52 & 11.79 \\
    \midrule
    GPFS & 349.12 & 6.37 & 53.91 \\
    SPFS & 573.32 & 6.30 & 90.69 \\
    GPIS & 422.79 & 6.37 & 65.74 \\
    SPIS & 600.88 & 6.23 & 96.21 \\
    \midrule
    GPFS-s & 229.29 & 9.71 & 22.61 \\
    SPFS-s & 213.83 & 9.30 & 21.99 \\
    GPIS-s & 232.20 & 9.85 & 22.56 \\
    SPIS-s & 217.19 & 9.58 & 21.67 \\
    \midrule
    PCFS-s & 189.57 & 8.84 & 20.44 \\
    PCIS-s & 152.9 & 7.31 & 19.93 \\

    \bottomrule
    \end{tabular}
    \label{tab:statistics}
\end{table}

\begin{table*}[tb]
    \centering
    \caption{LLM average evaluation scores on Persona Generated Datasets (1-5), the different aspects scored are as follows Q1(topic), Q2(situation), Q3(Qualities), Q4(Speech style), Q5(Age), Q6(lifeStyle), Q7(Memory), Q8 (Relations) and Q9(Overall) }
    \label{tab:GenGrounding}
    \begin{tabular}{L{1.5cm}R{1cm}R{1cm}R{1cm}R{1cm}R{1cm}R{1cm}R{1cm}R{1cm}R{1cm}R{1cm}}
    \toprule
    Dataset & Q1 & Q2 & Q3 & Q4 & Q5 & Q6 & Q7 & Q8 & Q9 & avgScore \\
    \midrule
    GPFS & 4.97 & 4.96 & 4.88 & 4.97 & 4.76 & 4.79 & 4.19 & 4.81 & 4.99 & 4.81 \\
    GPIS & 4.98 & 4.93 & 4.90 & 4.97 & 4.85 & 4.87 & 4.17 & 4.89 & 4.99 & 4.84 \\
    GPFS-s & 4.97 & 4.90 & 4.85 & 4.98 & 4.80 & 4.88 & 4.38 & 4.84 & 5.00 & 4.85 \\
    GPIS-s & 4.96 & 4.86 & 4.85 & 4.96 & 4.83 & 4.89 & 4.31 & 4.89 & 5.00 & 4.84 \\
    \bottomrule
    \end{tabular}
\end{table*}

\subsection{Measure of Lexical Diversity (MTLD)}
Following the methodology outlined in \cite{herbold2023large}, we employ the Measure of Lexical Diversity (MTLD) \cite{McCarthy2010MTLDVA} to compare human conversational datasets with our generated conversational datasets. According to the authors \cite{McCarthy2010MTLDVA}, MTLD is calculated as the mean length of sequential word strings in a text that maintain a given Type-Token Ratio value (TTR) i.e. percentage of unique tokens, by default set to 72\%. Our goal is to compare how much diversity is attained in conversations generated by LLM-based agents compared to casual human conversations. Table \ref{tab:mtld} presents the average MTLD score per conversation along with the standard deviation. It is observed that casual human conversational datasets generally exhibit lower MTLD scores compared to generated conversational datasets, which may be attributed to the nature of the casual conversation setting. Large Language Models (LLMs) inherently generate diverse vocabulary and use language in varied ways.
In the initial experiments, we achieve very high MTLD scores, comparable to those attained in academic writing style by humans. While a higher degree of diversity in conversations is desired, excessive diversity may be unusual for casual conversations. In experiments where agents are adjusted to be less verbose, the resulting conversations are shorter and exhibit relatively lower diversity, indicating more natural, yet still diverse, conversations.


\subsection{Effect of Iterative Sampling}
Iterative sampling was introduced to alleviate the tendency of LLMs to generate highly similar templatic conversations when presented with the same set of shots in large-scale data set generation. To estimate the effect of iterative sampling on the diversity of the generated conversations, we conduct PERSONA-CHAT-based experiments, primarily \textbf{PCFS-s} and \textbf{PCIS-s}. In these experiments, we fix the personas matched together based on the matching of interlocutors within the original PERSONA-CHAT dataset, particulary the test subset, thus the only change is disabling or enabling \textbf{iterative sampling} respectively. Table \ref{tab:mtld} demonstrates that using iterative sampling results in a slightly higher mean MTLD score (84.38) compared to fixing the shot during experience generation (80.11). This indicates that dynamically changing the shot using shots generated in previous runs can lead to more diverse content compared to using a single fixed experience shot.

\begin{table}[htbp]
    \centering
    \caption{Mean and Standard deviation of Measure of Lexical Diversity (MTLD) scores per conversation}
    \label{tab:mtld}
    \begin{tabular}{L{3cm}R{2cm}R{2cm}}
    \toprule
    Dataset & Mean MTLD & std MTLD \\
    \midrule
    DailyDialog & 53.44 & 19.54 \\
    EmpatheticDialogues & 63.47 & 19.47 \\
    PERSONA-CHAT & 49.58 & 12.29 \\
    \midrule
    GPFS & 107.24 & 16.67 \\
    SPFS & 125.16 & 20.02 \\
    GPIS & 111.47 & 16.28 \\
    SPIS & 129.11 & 19.28 \\
    \midrule
    GPFS-s & 86.87 & 17.15 \\
    SPFS-s & 98.53 & 27.11 \\
    GPIS-s & 85.95 & 16.79 \\
    SPIS-s & 101.76 & 28.79 \\
    \midrule
    PCFS-s & 80.11 & 20.06 \\
    PCIS-s & 84.38 & 19.24 \\
    \bottomrule
\end{tabular}
\end{table}

\subsection{LLM Evaluation}
We employ a methodology akin to that presented in the G-Eval framework \cite{Liu2023GEvalNE}, with the objective of achieving final scores that exhibit a high correlation with human evaluations. We establish a system using an LLM-as-a-judge approach, utilizing the \textbf{GPT-4o mini} model, henceforth referred to as the \textbf{judge}, to assess the alignment of generated conversations with the input experience provided by the experience generator.
To this end, we conduct multiple QA sessions with the judge for each input experience and conversation pair. During these sessions, we present two sequentially dependent questions to the judge. The initial question requires the judge to elucidate why it considers the conversation grounded in a particular aspect of the input experience. This step primarily aims to enrich the contextual understanding to enable the judge to provide a more representative score. Subsequently, based on the explanation given, the judge is asked to assign a score ranging from 1 to 5.
Consistent with the G-Eval framework, we utilize the returned log probabilities to normalize the output scores. The final score is computed as the weighted sum of each possible score (i.e., 1, 2, 3, 4, 5), each multiplied by the respective returned log probability for that token.

In experiments where personas are generated by the experience generator, a total of 9 questions are asked. These questions cover various aspects of groundedness, such as the groundedness of the conversation in the generated \textbf{topic (Q1)} and \textbf{situation (Q2)}, as well as its groundedness in the generated persona aspects such as \textbf{qualities (Q3)}, \textbf{speech style (Q4)}, \textbf{age (Q5)}, \textbf{lifestyle (Q6)}, \textbf{memories (Q7)}, and \textbf{relations (Q8)}. A final question is then used to assess the groundedness in the \textbf{overall generated experience (Q9)}. In experiments where personas are sampled, 4 questions are asked to evaluate the groundedness in the \textbf{topic (Q1)}, \textbf{situation (Q2)}, \textbf{personas of the interlocutors (Q3)}, and the \textbf{overall experience (Q4)}. Table \ref{tab:GenGrounding} and Table \ref{tab:SampleGrounding} presents the average score per dataset for each question, along with an overall average on the generated and sampled personas experiments respectively. The results in both tables indicate that the generated conversations are well-grounded in the generated experiences, and hence validating the usefulness of our framework in generating conversations guided by tuning the agents' behaviours.

\begin{table}[tb]
    \centering
    \caption{LLM average evaluation scores on Persona Sampled Datasets (1-5), the different aspects scored are as followed: Q1(topic), Q2(situation), Q3(Persona) and Q4(Overall)}
    \label{tab:SampleGrounding}
    \begin{tabular}{L{1.5cm}R{1cm}R{1cm}R{1cm}R{1cm}R{1cm}}
    \toprule
    Dataset & Q1 & Q2 & Q3 & Q4 & avgScore \\
    \midrule
    SPFS & 4.95 & 4.75 & 4.83 & 4.97 & 4.87 \\
    SPIS & 4.95 & 4.78 & 4.90 & 4.97 & 4.90 \\
    SPFS-s & 4.81 & 4.64 & 4.80 & 4.84 & 4.77 \\
    SPIS-s & 4.92 & 4.73 & 4.78 & 4.93 & 4.84 \\
    PCFS-s & 4.91 & 4.63 & 4.52 & 4.93 & 4.75 \\
    PCIS-s & 4.88 & 4.63 & 4.53 & 4.95 & 4.75 \\
    \bottomrule
    \end{tabular}
\end{table}

\section{Conclusion}
In this work, ConvoGen, a framework based on the Autogen framework, is introduced for generating open-domain conversational datasets. Various experiments with different configurations are conducted, demonstrating the impact of generating, sampling, and reusing personas on the conversations produced. The iterative sampling technique is proposed to increase the diversity of the generated content and avoid repetitive, templated conversations in large scale dataset generation. The generated conversations are evaluated in terms of their lexical diversity and groundedness in input experiences and compared to human-based casual conversational datasets. Our experiments show that ConvoGen is capable of generating diverse and well-grounded conversations in the input topic, personas and situation, and hence can be utilized for augmenting existing datasets for multi-party open-domain scenarios. However, it's crucial to acknowledge that content generated by LLMs can sometimes amplify harmful or explicit material and carry inherent biases, necessitating the utilization of content safety filters. Additionally, multi-agent frameworks like AutoGen may exhibit unpredictable behaviors, which can be mitigated through prompt tuning and conversation filtering to ensure the generated data's reliability.
\section*{Acknowledgment}

We used Microsoft Copilot for enhancing the writing of the sections in the paper.

\bibliographystyle{unsrt}
\bibliography{references}
\vspace{-2mm}
\clearpage
\onecolumn

\begin{appendices}

\section{Generation prompts}\label{appendix:genprompts}
\setcounter{figure}{0}
\subsection{Experience Generator Prompts}

\begin{figure}[h!]
    \centering
\begin{tcolorbox}[title=Experience Generator Prompt using Few-Shot Learning]
\small
You are a database helper agent who keeps track of records that document the relations between different individuals.
Your task is to present the records of those individuals and their relations between each other and imagine a situation that would involve all these individuals. 

\section*{Guidelines}
\begin{enumerate}
    \item First, decide on the number of individuals (2, 3, 4, or 5) and the nationality of those individuals and give each individual a name
    \item Decide on the qualities and personality of each individual: friendly, funny, ambitious, confident, caring, supportive, usually interrupts others
    \item Decide on the profession of each speaker: software engineer, baker, house wife,...
    \item Decide on the life style and hobbies for each speaker
    \item Decide on their short-term and long-term memory
    \item Decide on how each individual speaks (speech style)
    \item Decide on the relation of each individual to the other: friend, co-worker, mum, dad, son, daughter, manager,...
    \item Decide the situation or the setting which gathers these individuals. The situation can be either related to the long-term memory, or the short-term memory of one of the individuals, or a generic situation related to the profession or hobbies on of the speakers. The situation may also be a chit-chat about a general topic or a typical situation in the day of one of the speakers.
    \item Decide on the topic of the conversation related to the situation, the life-styles and the long and short term memories of at least one of the agents. 
    \item Provide a conversation starter for the topic. The conversation starter must be on a topic that could involve promises or generic chit-chat.
\end{enumerate}

\section*{Output Format}

Your output format is a json array where each json object is formatted as the following example:

\subsection*{Example}
```
json

\texttt{\{shots\}}

```

\subsection*{Output}

I will output an array of \texttt{\{numGeneratedExamples\}} json objects that follow the same format that is presented in output format. I \textbf{must never} copy the example as is, but use it as a reference in my generation.
I must make sure that I output valid json that must be parsed correctly.
size of the generated array = \texttt{\{numGeneratedExamples\}}
```json

\end{tcolorbox}
    \caption{Method 1: Prompt for generating experiences using few-shot learning}
    \label{prompt:experiencegen1}
\end{figure}

\begin{figure}[h!]
    \centering
\begin{tcolorbox}[title=Experience Generator Prompt using Few-Shot Learning - Sampled Personas]

\small
You are a database helper agent who keeps track of records that document the relations between different individuals.
You know the personas of these individuals and your **task** is to find the relations between these individuals and imagine a situation that would involve all these individuals

\section*{Guidelines}
\begin{enumerate}
\item First, Give each individual a real name and decide on the relation of each individual to the other: friend, co-worker, mum, dad, son, daughter, manager,...Names must not have spaces or digits.
\item Decide the situation or the setting which gathers these individuals. The situation can be based on a generic situation related to the profession or hobbies on of the speakers. The situation may also be a chit-chat about a general topic or a typical situation in the day of one of the speakers.
\item Decide on the topic of the conversation related to the situation, and the life-styles of at least one of the agents. 
\item Provide a conversation starter for the topic. The conversation starter must be on a topic that could involve promises or generic chit-chat.
\end{enumerate}
\section*{Output Format}

Your output format is a json array where each json object is formatted as the following example:

\subsection*{Example}

```
json

\texttt{\{shots\}}

```

\subsection*{Input}

Your input is an array of\texttt{\{numGeneratedExamples\}} groups. Each group comprises an array of individual personas

\subsection*{Output}

I will output an array of \texttt{\{numGeneratedExamples\}} json objects that follow the same format that is presented in output format example.
Each element in the output array represents the relations between the corresponding group of individuals in the input array, as well as the situation involving these individuals, a topic for their conversation and a conversation starter.
I **must never** copy the example as is, but use it as a reference in my generation.
I must make sure that I output valid json that must be parsed correctly.
size of the generated array = \texttt{\{numGeneratedExamples\}}

Input:

\texttt{\{personas\}}

```json

\end{tcolorbox}
    \caption{Method 2: Prompt for generating experiences using few-shot learning and input personas sampled from the persona hub}
    \label{prompt:experiencegenpersona}
\end{figure}

\clearpage
\subsection{Agent Prompts}

\begin{figure}[h!]
    \centering
    \begin{tcolorbox}[title=Agent Conversation Guidelines]

\subsection*{Guidelines for the conversation:}
\begin{itemize}
\item You don't need to address the other speakers by their names. You don't need to speak to everyone in the conversation. For example you can say "I think that's a great idea." instead of "I think that's a great idea, John.". This makes the conversation more natural.
\item You can ask questions, provide answers, and make comments.
\item You can also provide information and share your opinions.
\item Never sound artificial or robotic. For example, instead of saying "Your project sounds fascinating, Lucas.", you can say "fascinating, yeah".
\item You can stop the conversation at any time by saying "I have to go now." or "I have to leave now."
\item You can pause in the middle of the conversation by saying "I need a ", to allow other speakers to interrupt you.
\item You can interrupt other speakers by saying "I have something to say." or "I have a question."
\item You can express your promises by saying "I will do that." or "I promise to do that."
\item You don't need to start with confirmation words like "yes" or "okay" or "Absolutely". You can start with the main content of your response, so that you can sound more natural.
\item **You must be concise and limit your response to 30 tokens at most.**
\end{itemize}

    \end{tcolorbox}
    \caption{Guidelines for each agent to drive the conversation}
    \label{fig:agent-guidelines}
\end{figure}
\end{appendices}

\end{document}